\documentclass{article}

\usepackage{main}

\usepackage[utf8]{inputenc} 
\usepackage[T1]{fontenc}    
\usepackage{hyperref}       
\usepackage{url}            
\usepackage{booktabs}       
\usepackage{amsfonts}       
\usepackage{nicefrac}       
\usepackage{microtype}      
\usepackage{xcolor}         
\usepackage{graphicx}
\usepackage{caption}
\usepackage{subcaption}
\usepackage{algorithm}
\usepackage{algorithmic}
\usepackage{float}
\usepackage{wrapfig}
\usepackage{titlesec}
\usepackage{amsmath}
\titlespacing*{\section}{0pt}{1.1\baselineskip}{\baselineskip}
\title{Safe Evaluation For Offline Learning: \\Are We Ready To Deploy?}

\author{
    Hager Radi \\
  University of Alberta \\
  Alberta, Canada \\
  \texttt{radi@ualberta.ca} \\
   \And
   Josiah P. Hanna \\
   University of Wisconsin -- Madison \\
   Wisconsin, USA\\
   \texttt{jphanna@cs.wisc.edu} \\
   \AND
   Peter Stone \\
   The University of Texas at Austin \& \\
   Sony AI \\
   Texas, USA \\
   \texttt{pstone@cs.utexas.edu} \\
   \And
   Matthew E. Taylor \\
   University of Alberta \& \\ Alberta Machine Intelligence Institute\\
   Alberta, Canada \\
   \texttt{matthew.e.taylor@ualberta.ca} \\
}

\begin{document}

\maketitle

\begin{abstract}
    The world currently offers an abundance of data in multiple domains, from which we can learn reinforcement learning (RL) policies without further interaction with the environment. RL agents learning offline from such data is possible but deploying them while learning might be dangerous in domains where safety is critical. Therefore, it is essential to find a way to estimate how a newly-learned agent will perform if deployed in the target environment before actually deploying it and without the risk of overestimating its true performance.
    To achieve this, we introduce a framework for safe evaluation of offline learning using approximate high-confidence off-policy evaluation (HCOPE) to estimate the performance of offline policies during learning. In our setting, we assume a source of data, which we split into a train-set, to learn an offline policy, and a test-set, to estimate a lower-bound on the offline policy using off-policy evaluation with bootstrapping. A lower-bound estimate tells us how good a newly-learned target policy would perform before it is deployed in the real environment, and therefore allows us to decide when to deploy our learned policy.
\end{abstract}

\section{Introduction}
\begin{wrapfigure}{r}{0.4\linewidth}
\vspace{-0.25cm}
    \includegraphics[width=0.9\linewidth, height=3cm]{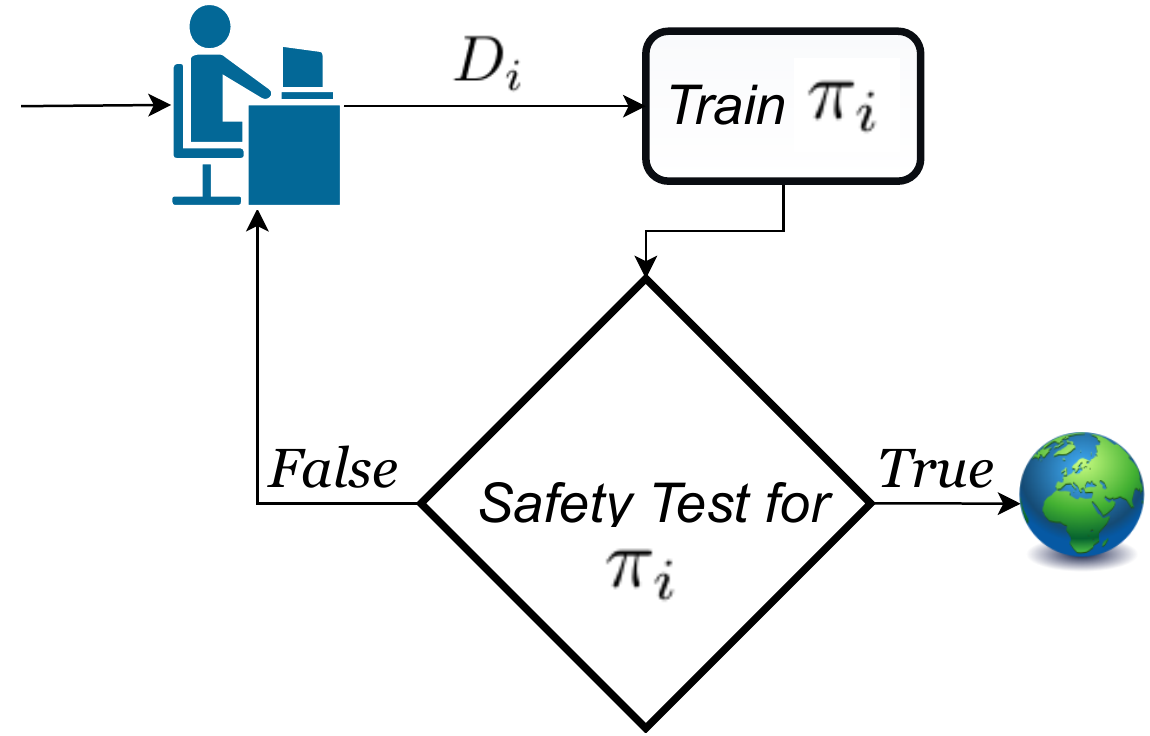}\hfil
    \caption{A framework for continual safety-evaluation of offline learning}
    \label{setup}
    \vspace{-0.2cm}
\end{wrapfigure}
Suppose someone else is controlling a sequential decision making task for you. This could be a person trading stocks for you, a hand-coded controller for a chemical plant, or even a PID for temperature regulation. Off-policy reinforcement learning allows us to learn policies from fixed data. But when would you want to switch from the existing controller to your learned policy? This decision may depend on the cost for continued data collection from the existing controller (e.g., you pay someone to manage your stock portfolio), your risk appetite, and your confidence in the performance of the policy you have learned. This paper takes a critical first step towards the last question: how to learn off-policy and simultaneously evaluate that policy with confidence when we have no access to the environment nor the policy generating the data?

Offline reinforcement learning is a way of training off-policy algorithms offline using already existing data, generated by humans or other controllers. Learning with offline data is challenging because of the distribution mismatch between data collected by the behavior agent and the offline agent \cite{LevineOfflineRLtutorial}. What is even more challenging is evaluating offline agents in the offline RL setting if we assume no access to the environment; in some domains, we cannot execute our learned policy until it is good enough. This limitation raises the question about the possibility of using off-policy policy evaluation (OPE) methods, where we can estimate the value of a policy using trajectories from another policy, to predict what the performance of the offline agent is, at any point of time during learning. Further, we investigate if we can use high-confidence OPE (HCOPE) to control the risk of overestimating the policy's performance.

We present a framework for safe evaluation of RL agents learning offline. \textit{Safe} refers to defining a lower bound to our policy estimates, a point where we think the new policy is worse than it is in practice, which is important for safety-critical applications. The paper tackles the setup where we combine offline policy improvement to learn a policy from existing data, and off-policy policy evaluation with bootstrapping confidence intervals to provide a lower confidence bound estimate of such a policy, simultaneously. We believe the setup we are tackling is under-studied in the literature. We are aiming to learn a target policy purely out of a data buffer, and evaluate its performance while learning so that we can tell when it is ready for deployment. Our framework can be summarized in Figure \ref{setup}: we have a source of data, from which we utilize samples. At each step, we split the data into training and testing. After each training step, we test the policy using approximate \footnote{We refer to HCOPE as approximate because bootstrapping lower bounds may have error rates larger than $\delta$ but they provide a practical alternative to guaranteed bounds that are too loose to use.} high-confidence off-policy evaluation (HCOPE). We continue the process of training/testing for a few iterations until the testing shows the policy can outperform the data with a confidence level $\delta$. We dynamically receive samples, continuously performs RL updates, and continuously monitors a confidence interval on the changing policy until it reaches a sufficient level. Our \textbf{contributions} are:
\begin{itemize}
    \item A framework combining offline RL and approximate high-confidence OPE.
    \item Investigating the feasibility of HCOPE methods given constantly-improving target policy (as opposed to a fixed policy previously studied in the literature)
\end{itemize}





\section{Motivation}
Offline RL offers an opportunity for learning data-driven policies without environment interaction. In safety-critical applications found in healthcare or autonomous driving, there are plenty of data that we can use to learn RL policies and hence use for decision making \cite{guidelines_RL_nature}.
Also, there are domains where data efficiency is essential as the data collection process is either expensive or dangerous. If we want to learn policies in such domains, we need to find a way to tell how good the performance of a policy is before actually deploying it to the real-world environment. However, the execution of a new policy can be costly or dangerous if it performs worse than the policy that is currently being used. Hence, we focus on safety when evaluating offline agents such that the probability that the performance of our agent below a baseline is at most $\delta$, where $\delta$ specifies how much risk is reasonable for a certain domain. This will allow us to know when we can trust a policy to take control in the real world. If we are given access to trajectories generated by an unknown behavior policy, we assume an iterative setting where at each iteration we can either request another batch of trajectories from the source or deploy our own policy. Our objective is to only deploy our own policy if a $1-\delta$ lower bound on its expected return is greater than the expected return of the unknown behavior policy.
\section{Related work} \label{related_work}
The current work intersects with the literature of two camps, which are \textit{offline RL} and \textit{batch RL}. \textit{Offline RL} is about improving a policy from historical data for control, not just evaluation. A survey paper \cite{LevineOfflineRLtutorial} discusses how to categorize model-free offline RL methods into policy constraints that constrain the learned policy to be close to the behavior policy such as the work by \citet{bcq}, and uncertainty-based methods that attempt to estimate the epistemic uncertainty of Q-values to reduce distributional shift. Non-constrained methods include the traditional Q-learning \cite{Q_learning}, Double DQN \citet{double_dqn}, or Soft-Actor Critic \cite{SAC} but they are not always reliable in the fully offline setup. Imitation learning, more specifically behavioral cloning \cite{bc}, is another way for learning offline policies from historical data. 

\textit{Batch RL} refers to learning off-policy estimates from historical data, without environment interaction. It has an important property: given data generated by a behavior policy, it will estimate a new evaluation/target policy, guaranteed with high confidence that its performance is not worse than the behavior policy $\pi_b$. This camp includes three sub-directions. Off-policy policy evaluation \textit{(OPE)} \cite{empirical_ope}, high confidence off-policy policy evaluation \textit{(HCOPE)} \cite{Thomas2015HighConfidenceOPE}, and safe policy improvement \textit{(SPI)} \cite{Thomas2015HighConfidencePolicyImprovement}. An empirical study of OPE methods \cite{empirical_ope} discussed the applicability of each method and presented method selection guidelines depending on the environment parameters and the mismatch between $\pi_\theta$ and $\pi_b$. This study categorized OPE methods into importance sampling methods, directed methods, and hybrid methods that combine aspects from the two worlds. Importance Sampling (IS) \cite{importance_sampling} is one of the widely used methods for off-policy evaluation where rewards are re-weighted by the ratio between $\pi_\theta$ and $\pi_b$. There are later versions such as Weighted Importance Sampling \cite{WIS}, Per-Decision Importance Sampling, and Per-Decision Weighted Importance Sampling \cite{importance_sampling} that are biased, but offer lower-variance estimates. Then, we have the direct methods which rely on regression techniques to directly estimate the value function of the policy. This category includes model-based methods where the transition dynamics and reward are estimated from historical data via a model. Then, the off-policy estimate is computed with Monte-Carlo policy evaluation \cite{Hanna_MB}. Another direct method is Fitted Q-Evaluation \cite{fitted_q_evaluation}, which is a model-free approach and acts as the policy evaluation counter-part to batch Q-learning or FQI \cite{FQI}. The third category is the hybrid methods that combine different features from IS and direct methods. This category mainly involves doubly-robust methods \cite{dr_methods} that uses a direct model to lower the variance of IS.

Given the previous work, none of them discussed the feasibility of evaluating offline RL agents during learning, and how much we can trust high-confidence OPE as an approach for testing. The setup we are studying is quite different from the current literature because previous work assumed a behavior policy $\pi_b$ and a target policy $\pi_\theta$, where both policies are fixed and may be related. As an example, in a study for HCOPE methods by \citet{Thomas2015HighConfidenceOPE}, the target policy is initialised as a subset of the behavior policy such that they are close to each other. In another study for safe improvement \cite{Thomas2015HighConfidencePolicyImprovement}, the Daedulus algorithm learns a safe target policy as a continuous improvement over the behavior policy. This is quite relevant to what we have here but Daedulus uses data from an older version of the policy it currently improves to perform the improvement step and safety tests. With the Doubly-robust estimator, authors present the results with different versions of $\pi_\theta$ such that $\pi_\theta$ is always a mixture of $\pi_b$ and $\pi_\theta$ with different degrees \cite{dr_methods}. To the best of our knowledge, none of the previous work showed how OPE or HCOPE methods would perform if the target policy is improved independently from the behavior policy, which is the case for offline learning.

\section{Methodology}
In this work, we present a framework to close the gap between offline RL and approximate high-confidence off-policy evaluation as a feasible evaluation method for offline agents in safety-critical domains. In our framework, we sample data dynamically at each iteration, perform policy updates, and continuously monitor its performance with a confidence interval till it reaches a good performance and is ready to be deployed. The policy we are learning offline does not interact with the environment unless it passes the safety test. Specifically, we have $k$ iterations where in each iteration $i$, the current data size $n$ is denoted as $D_i$. $D_i$ is split between training our target policy $\pi_{\theta_i}$ and performing the safety test using approximate high-confidence off-policy evaluation methods. We return when the learned policy is ready to be deployed with appropriate confidence.
To summarize an interaction between an agent and a data source: 1) an agent requests a set of data $D_i$ without knowledge of the behavior policy, 2) we improve a policy $\pi_{\theta_i}$ offline using $D_{train}$ with any offline RL algorithm, 3) we perform a safety check using $D_{test}$ with high-confidence evaluation methods, and 4) our new policy is either ready to be deployed or we go back to step 1). \\
Instantiating our framework requires selecting an offline policy improvement method and a method for computing off-policy lower-bound estimates. For these two components, we study a variety of methods. We refer to each offline policy improvement method with $\Phi$, and high-confidence off-policy evaluation method with $\Psi$. Our safe evaluation framework is further explained in Algorithm \ref{alg:algorithm_1}.
\begin{algorithm}[tb]
\caption{Offline Safe Evaluation Framework}
\label{alg:algorithm_1}
\textbf{Input}: initial $\pi_\theta$, dataset $D$ of $n$ trajectories, confidence level $\delta \in [0, 1]$, number of bootstrap estimates $B$, policy improvement method $\Phi$, HCOPE method $\Psi$\\
\textbf{Output}: $\pi_\theta$, $\hat{v_\delta}(\pi_\theta)$: $1-\delta$ lower-bound on $v(\pi_\theta)$
\begin{algorithmic}[1] 
\STATE Let $i=0$
\WHILE{  $\hat{v_\delta}(\pi_\theta) \leq \hat{v}(\pi_b)$}
\STATE Request data-set of trajectories $D_i$ with size $n$
\STATE Split $D_{i}$ into $D_{train}$ and $D_{test}$
\STATE Improve policy $\pi_{\theta_i} = \Phi(D_{train})$
\STATE Evaluate policy $\hat{v_\delta}(\pi_\theta) = \Psi (D_{test})$
\ENDWHILE
\STATE \textbf{return} $\pi_\theta$, $\hat{v_\delta}(\pi_\theta)$
\end{algorithmic}
\end{algorithm}
\subsection{Policy improvement}
We investigate different offline learning techniques such as naive off-policy methods (Category A), imitation learning (Category B), and policy constraint approaches specific for offline learning (Category C). We use Double DQN \cite{double_dqn} in Category A, Behavioral Cloning (BC) in Category B, and Batched Constrained Q-Learning (BCQ) \cite{benchmarking_batchRL} in Category C. 
\subsection{Approximate high-confidence off-policy evaluation with bootstrapping}
We use off-policy estimators to evaluate a policy that is continuously learning. An off-policy estimator is a method for computing an estimate ${\hat{v}(\pi_\theta)}$ on the true value of the target policy ${v(\pi_\theta)}$ using trajectories $D$ from another policy $\pi_b$, which is what Off-policy policy evaluation (OPE) methods do. For high-confidence OPE, this extends to lower-bounding the performance of the target policy $\pi_\theta$ where we combine OPE with bootstrapping confidence intervals. Bootstrapping is considered semi-safe since it requires the assumption that the bootstrap distribution can represent the statistic of interest which is not true for a finite sample, which in result makes our HCOPE approximate. However, it is safe enough as well as data efficient for high-risk domains like medical predictions \cite{bootstrap_for_healthcare}. When evaluating, we use new samples each time, to avoid the multiple comparisons problem\footnote{The problem occurs when conducting multiple statistical tests simultaneously with reusing data so the confidence bound does not strictly hold anymore}. This also ensures that we do not overfit our OPE estimates or tune training parameters to reduce the estimate error. The evaluation approximates a confidence lower bound of $\hat{v_\delta}(\pi_\theta)$ on $v(\pi_\theta)$ such that $\hat{v_\delta}(\pi_\theta) \leq v(\pi_\theta)$ with probability at least $1-\delta$. There are 3 main categories of estimators as mentioned in Section \ref{related_work}; we use weighted importance sampling with bootstrapping (WIS), direct model-based bootstrapping (MB), and weighted doubly-robust with bootstrapping (WDR) to represent each of the 3 categories of off-policy estimators. Further details can be found in Appendix \ref{app_1}.
\section{Experiments}
In our experiments, we focus on discrete control in the classic MountainCar \cite{sutton} with a shortened horizon\footnote{Further details in Appendix \ref{app_2}}. To mimic the setup of Figure \ref{setup}, we simulate a data source with an online partially-trained actor-critic that collects data of a medium quality. To include exploratory behavior in data, an agent takes a random action $30\%$ of the time instead of following the online policy when collecting trajectories. We improve a policy offline using 3 different improvement methods (BC, BCQ, and Double DQN), and evaluate each policy simultaneously using $3$ different off-policy evaluation methods (WIS, MB, and WDR). We follow Algorithm \ref{alg:algorithm_1}, but we limit the loop to maximum iterations of $10$ because not all evaluation methods will be able to achieve the stopping condition. In each iteration, we sample $300$ trajectories from the data source and split between training and testing such that training uses $20$ trajectories only since evaluation requires more data for tighter bounds. For each improvement method, we pass training data and do $m$ policy updates where $m$ is tuned per algorithm such that the offline policy converges within the total number of iterations. For bootstrapping, we use ($\delta = 0.05$) to get a $95\%$ confidence lower-bound using $B=2000$ bootstrap estimates as recommended by practitioners \cite{efron_bootstrap}. For WDR only, we use a value of $B=224$ to avoid the heavy computation; this can still get us a good approximation as suggested by MAGIC \cite{WDR_and_Magic}.
\begin{figure}
\begin{subfigure}{.33\textwidth}
  \centering
  \includegraphics[width=\textwidth,height=4cm]{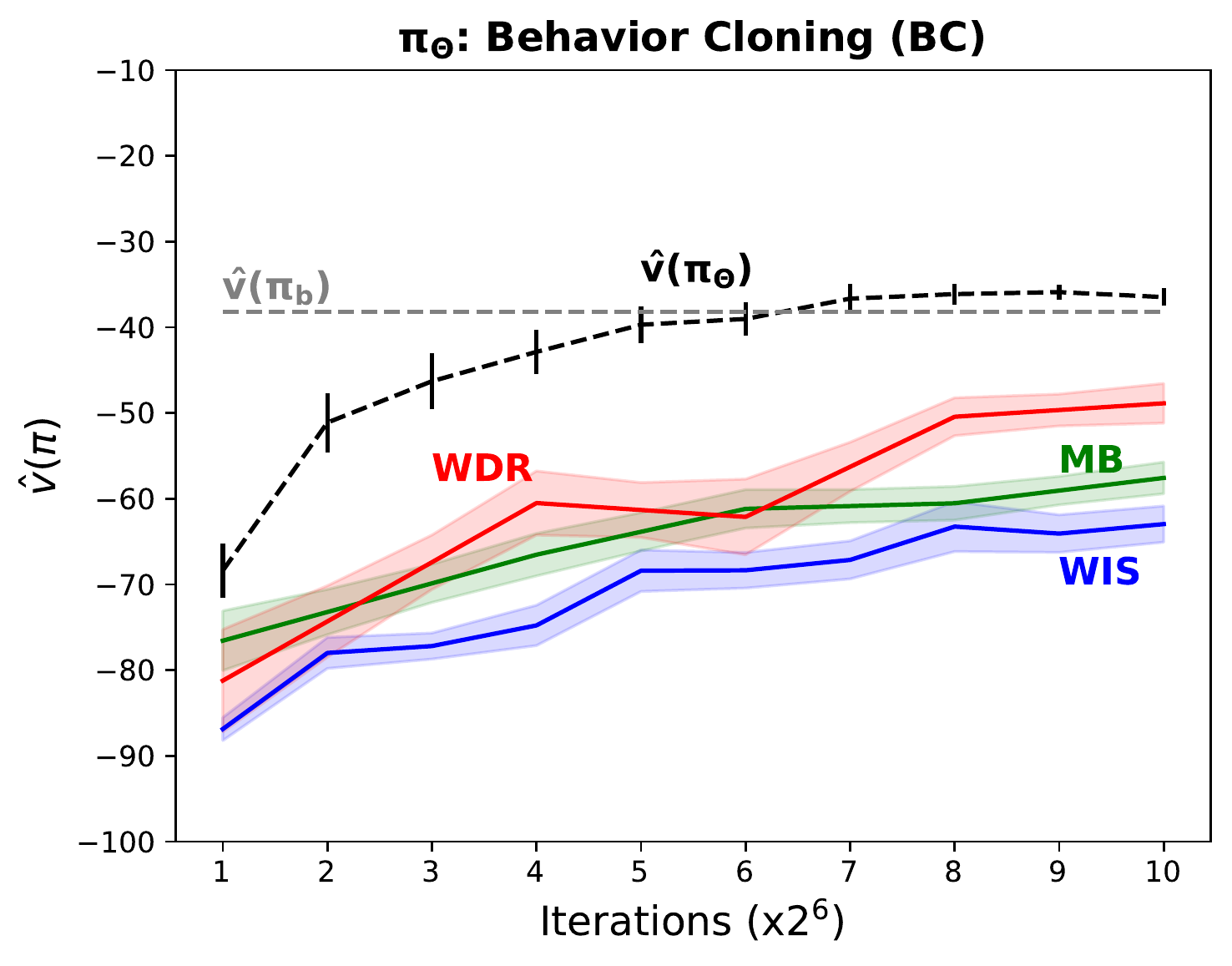}
\end{subfigure}%
\begin{subfigure}{.33\textwidth}
  \centering
  \includegraphics[width=\textwidth,height=4cm]{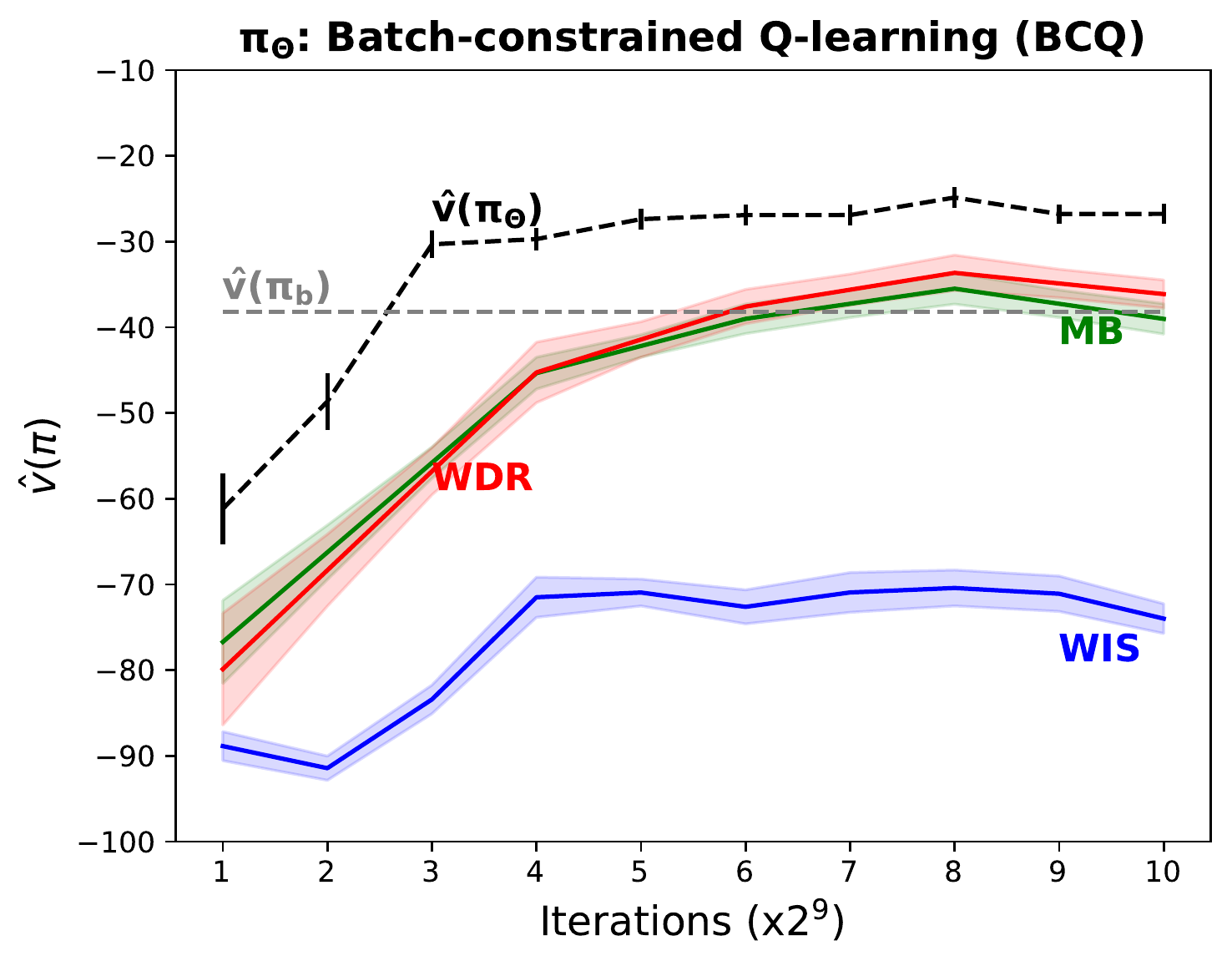}
\end{subfigure}
\begin{subfigure}{.33\textwidth}
  \centering
  \includegraphics[width=\textwidth,height=4cm]{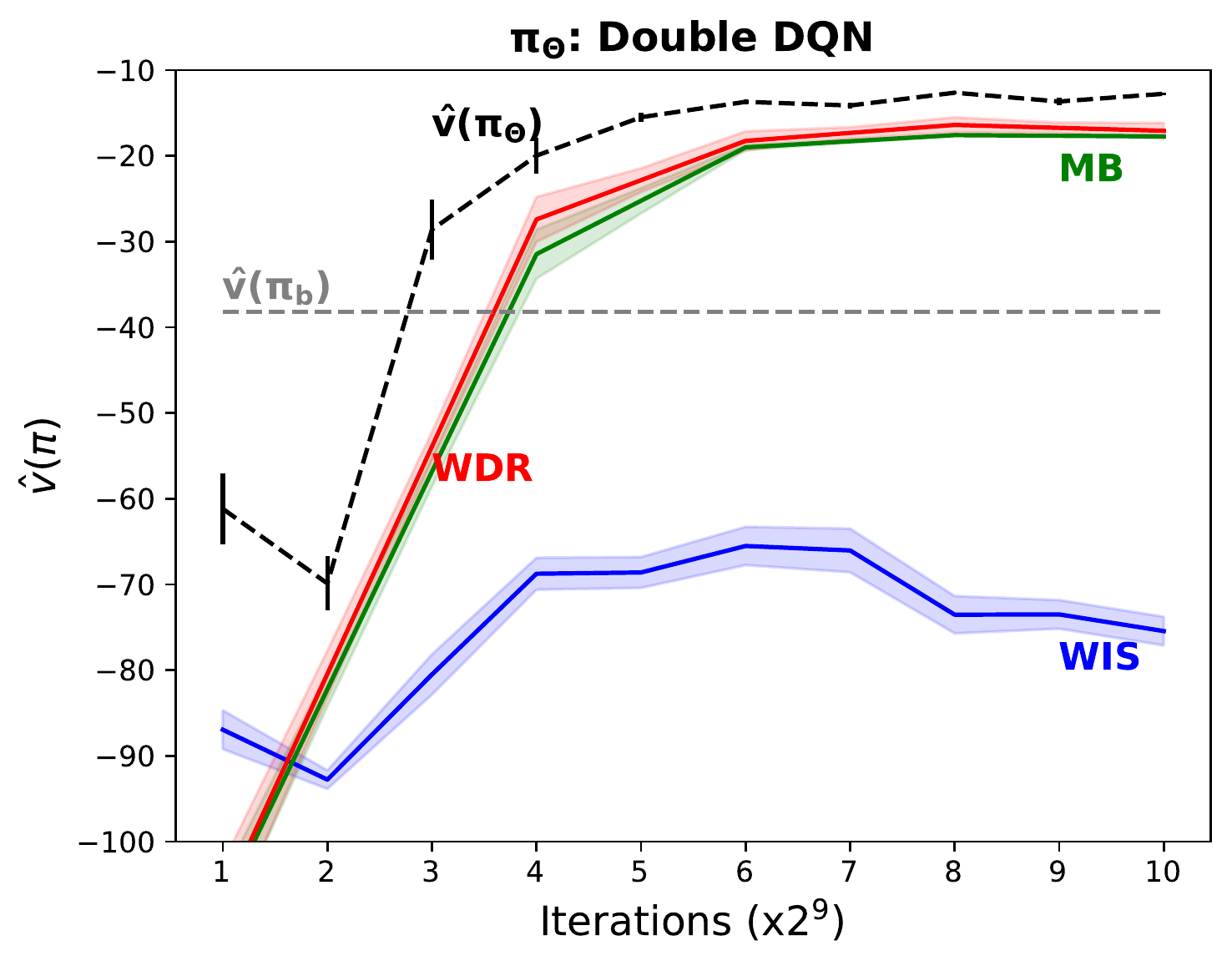}
\end{subfigure}%
\caption{\small Results of safe evaluation using different offline improvement algorithms on MountainCar. Double DQN and BCQ outperforms the data distribution while BC reaches $\pi_b$ at most. WIS, MB, and WDR refer to the evaluation methods: weighted importance sampling, model-based, and weighted doubly-robust respectively.}
\label{fig:MC_results}
\end{figure}


\textbf{Results}: Figure \ref{fig:MC_results} shows how different offline policy improvement methods perform given medium-quality data along with HCOPE estimators, indicating which method can tell when $v(\pi_\theta) > v(\pi_b)$. Behavior cloning as an offline policy improvement method can only perform as well as the data by $\pi_b$ while Double DQN and BCQ were able to outperform $\pi_b$. The true value of a target policy is calculated as the average return when running the policy in the actual environment (not possible in practice) for $1000$ episodes. All reported results are average of 40 runs while the shaded area shows the standard error. The value of the behavioral policy $\hat{v}(\pi_b)$ is the sum of undiscounted rewards of the data set. Since $\pi_b$ is not known, we first estimate $\hat{\pi_b}$ given the test data \cite{bpolicy_estimation}.

For the safe evaluation of each offline agent, weighted importance sampling (WIS) with bootstrapping failed to detect that the offline policy outperforms $\pi_b$ for all improvement methods. The model-based estimator (MB) and weighted doubly-robust with bootstrapping (WDR) have much less error with the true value of a policy and were able to inform when an offline agent outperforms $\pi_b$ and hence ready to be deployed. 
For instance, with 95\% confidence, in case of offline improvement with Double DQN, we were able to tell that our new target policy is better than the behavior policy at iteration 4 using two estimators (MB and WDR). Our ability to detect an improved policy is a function of how much better that policy is and how good our OPE methods are. It is better to rely on estimators that do not take $\pi_b$ into account (e.g. direct methods or hybrid methods) so that estimates are less affected by the offline improvement method and its divergence from $\pi_b$ (as the case for WIS). Further analysis can be found in Appendix \ref{app_3}. Accordingly, high-confidence off-policy estimators (e.g. direct methods or hybrid methods) are a safe evaluation method for offline learning with enough confidence that controls the risk of overestimating the true performance of an offline policy. 
\section{Conclusion}
In this paper, we propose a framework for safe evaluation of offline RL methods. While dynamically receiving data, we train offline RL agents and run safety tests to estimate a lower-bound on the value of the target policy and control the risk of overestimating its true value. This is essential for safety-critical applications to be able to tell when it is safe to deploy a new policy. We believe safe evaluation is an important step for offline RL; offline RL has great potential for control in the actual environment, if they are good enough, where our proposed framework is applicable. In future work, we will test this framework on more complex environments, both discrete and continuous. We will include learning offline from multiple sources of data with different qualities and explore how the quality of the data affects safe evaluation. Moreover, we can use safety testing to develop algorithms which can estimate in advance how much data is needed to learn a good-enough policy that can take over the real environment with maximum data efficiency.

\begin{ack}
This work has taken place in the Intelligent Robot Learning (IRL) Lab at the University of Alberta, which is supported in part by research grants from the Alberta Machine Intelligence Institute
(Amii); a Canada CIFAR AI Chair, Amii; NSERC; and Compute Canada. A portion of this work has taken place in the Learning Agents Research Group (LARG) at UT Austin.  LARG research is supported in part by NSF (CPS-1739964, IIS-1724157, FAIN-2019844), ONR (N00014-18-2243), ARO(W911NF-19-2-0333), DARPA, Lockheed Martin, GM, Bosch, and UT Austin's Good Systems grand challenge.  Peter Stone serves as the Executive Director of Sony AI America and receives financial compensation for this work. The terms of this arrangement have been reviewed andapproved by the University of Texas at Austin in accordance with its policy on objectivity in research.

\end{ack}

\newpage
\small
\bibliography{main.bib}

\newpage
\appendix

\section{Appendix}
\subsection{Bootstrapping} \label{app_0}
Consider a sample $D$ of $n$ random variables $H_j$ for $j = 1, 2, ..., n$ where we can sample $H_j$ from some $i.i.d.$ distribution of data. From the sample of data $D$, we can compute a sample estimate $\hat{\theta}$ of a parameter $\theta$ such  that $\hat{\theta} = f(D)$ where $f$ is the function to compute $\theta$. Given a dataset $D$, we create $B$ resamples with replacement, where $B$ is the number of bootstrap resamples, and compute $\theta$, $\hat{\theta}$, on each of these resamples. Bootstrapping \cite{efron_bootstrap} allows us to estimate the distribution of $\hat{\theta}$ with confidence intervals. The estimates computed with different resamples will be used to determine the $1-\delta$ confidence interval. In our setup, the parameter of interest $\theta$ is the expected return of a policy $v(\pi)$.

Hence, with a confidence level $\delta \in [0, 1]$ and $B$ resamples of the dataset of trajectories $D$, we use bootstrapping methods to approximate a confidence lower bound of $v_\delta(\pi)$ on $v(\pi)$ such that $v_\delta(\pi) \leq v(\pi)$ with probability at least $1-\delta$. 

As the size of data $n \to \infty$, bootstrapping has strong guarantees but it lacks guarantees for finite samples as it is not an exact method. To use bootstrapping, we have to assume that the bootstrap distribution is representative of the distribution of the statistic of interest, which is not the case for finite samples \cite{Hanna_MB}. As a result, bootstrapping is considered semi-safe but it is still safe enough for high risk medical predictions in practice with a known record of producing accurate confidence intervals \cite{bootstrap_for_healthcare}.

    
\subsection{High-confidence off-policy evaluation techniques} \label{app_1}
\paragraph{Importance sampling with bootstrapping} Importance Sampling (IS) \cite{importance_sampling} is a way for handling mismatch between distributions and hence presented as a consistent and unbiased off-policy estimator. For a trajectory $H \sim \pi_b$ of length $L$, as $H = S_1, A_1,.., S_L, A_L$, we can define the importance sampling up to time $t$ for policy $\pi_\theta$ as follows:
\begin{equation*}
        IS(\pi_\theta, \pi_b, D) = \sum_{i=1}^m \rho_{L}^H R^i, \rho_t^H := \prod_{j=0}^t \frac{\pi_\theta(A_j | S_j)}{\pi_b(A_j | S_j)}
\end{equation*}
In our setup given no access to $\pi_b$, we estimate the behavior policy $\pi_b$ from data but it results in a biased estimator \cite{bpolicy_estimation}. To produce a high-confidence IS estimate, we first compute the importance weighted returns then use bootstrapping to get the lower-bound estimate. A popular method is Bias Corrected and accelerated (BCa) bootstrapping \cite{efron_bootstrap}. A pseudo code of BCa is well-described as Algorithm 3 in \cite{Thomas2015HighConfidencePolicyImprovement}. We use weighted importance sampling \cite{WIS} for discrete control along with bootstrapping.
\paragraph{Direct model-based with bootstrapping} Model-based estimation is another off-policy estimator that lies under the direct methods. The model-based off-policy estimator MB computes $v(\pi_\theta)$ by building a model using all the available trajectories $D$ to build a model $\hat{M} = (S, A, \hat{P}, r, \gamma, \hat{d_0})$ where $\hat{P}$ and $\hat{d_0}$ are estimated with trajectories sampled from $\pi_b$. Then, MB will compute $\hat{v}(\pi_\theta)$ as the average return of trajectories simulated in the estimated model $\hat{M}$ while following $\pi_\theta$. Despite having lower variance than IS methods, MB is an inconsistent estimator such that as $n \rightarrow \infty$, the model estimates may converge to a value different from $V(\pi_\theta)$. This is because it is dependant on the modeling assumptions we make, whether we assume a linear or a non-linear model. To get a lower confidence bound on the MB estimate, we use bootstrap confidence intervals; the exact algorithm used is detailed in \citet{Hanna_MB}.
This bootstrapping method is quite similar to what we mentioned in the previous paragraph but much simpler in implementation and adaptable to different OPE methods. We rely on the model-based estimator with bootstrapping in our study as one of OPE direct-methods.
\paragraph{Weighted doubly-robust with bootstrapping} Weighted Doubly-Robust (WDR) \cite{WDR_and_Magic} is a hybrid method for off-policy estimation, presented as an extension to the doubly-robust (DR) method \cite{dr_methods}. DR is an unbiased estimator of $v(\pi_\theta)$ that uses an approximate model of the MDP to reduce the variance of importance sampling \cite{dr_methods}.
Although biased, WDR is based on per-decision weighted importance sampling (PDWIS) and serves as an improvement over DR method as it significantly balances the bias-variance trade-off. The approximate model value functions act as a control variate for PDWIS \cite{WDR_and_Magic}. 
\begin{equation}
    PDWIS(\pi_\theta, D, \pi_b) = \sum_{i=1}^{m} \sum_{t=0}^{L-1} \frac{\rho_t^i}{\sum_{j=1}^m \rho_t^j} \gamma^t R_t^i
\end{equation}

\begin{equation}
\begin{split}
WDR(\pi_\theta, D, \pi_b) & = PDWIS(\pi_\theta, D, \pi_b) - \\ & \sum_{i=1}^{n} \sum_{t=0}^{L-1} \gamma^t (w_t^i \hat{q}_{\pi_\theta} (S_t^i, A_t^i) - w_{t-1}^i \hat{v}_{pi_\theta} (S_t^i))
\end{split} 
\end{equation}
In previous methods, we use bootstrapping with WDR to provide lower confidence bound estimates over $v(\pi_\theta)$. WDR with bootstrapping is guaranteed to converge to the correct estimate as $n$ increases given the statistical consistency of PDWIS \cite{Hanna_MB}. For the approximate model, a single model is estimated with the available trajectories $D$, and used to compute the value functions of WDR for each bootstrap data. We choose the weighted doubly robust estimator to represent OPE hybrid methods in our study.

\subsection{Experimental Details} \label{app_2}
\textbf{Environment}: We use the classic MountainCar \cite{sutton} with a continuous state (velocity and position), and 3 possible discrete actions. At each time-step, the reward is $-1$ except for in a terminal state when it is $0$. However, we used the modified version of MountainCar as described by \citet{phil_thesis}. This means we shorten the horizon of the problem by holding an action $a_{t}$ constant for $4$ updates of the environment state. We also change the start state such that an episode starts with a random position and random velocity \cite{dr_methods, phil_thesis}.

\subsection{Analysis of evaluation methods} \label{app_3}
To better understand how safety testing is affected by the policy improvement methods, there are multiple measures to tell the similarity between two probability distributions (as a policy is a distribution over states and actions). One of the practical measures is the total variation (TV) distance. TV distance is a way to measure the difference between action probabilities taken under two policies given the data set in hand. TV would be the sum of differences in probabilities between the behavior policy $\pi_b$ and the target policy $\pi_\theta$ for each state-action pair in the test set of data. Since $\pi_b$ is not known, we estimate $\hat{\pi_b}$ given the test data \cite{bpolicy_estimation}.

\begin{wrapfigure}{r}{0.6\linewidth}
\vspace{-0.25cm}
    \includegraphics[width=0.9\linewidth,height=5cm]{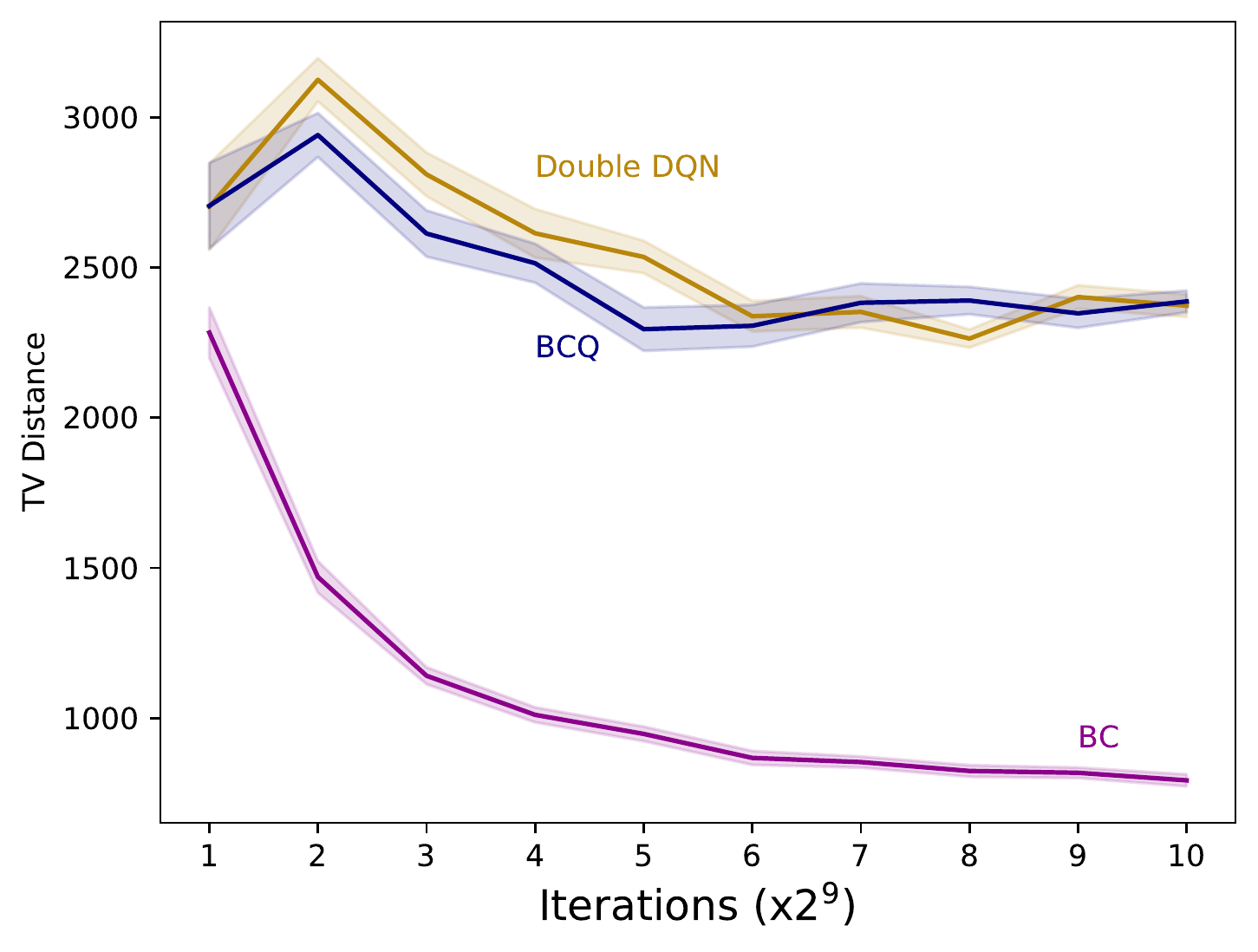}\hfil
    \caption{Total Variation Distance between $\hat{\pi}_b$ \& $\pi_\theta$}
    \label{tvd}
    \vspace{-0.cm}
\end{wrapfigure}

We analyze how the improvement method is affecting TV distance between $\hat{\pi_b}$ and $\pi_\theta$ and hence affecting the high-confidence off-policy evaluation estimates. TV distance is correlated to KL-Divergence and hence shows how two policies are different from each other. Figure \ref{tvd} shows the total variation distance between the offline policy $\pi_\theta$ and the estimated behavior policy $\hat{\pi}_b$ across the different offline learning methods. It show that behavioral cloning (imitation learning) can achieve a much lower distance than other improvement methods whether they are constrained or non-constrained (Double DQN and batch-constrained Q-learning); this is because behavioral cloning forces its target policy to be close to the behavior policy while other methods do not. This result also explains why weighted importance sampling achieves the lowest error between the estimate and the true value in the case of behavioral cloning; the error grows for other methods that do not constrain the policy to be close to the data distribution.


\end{document}